\documentclass{article}
\usepackage{spconf,amsmath,amssymb,graphicx,booktabs,microtype}
\usepackage[hidelinks]{hyperref}
\newcommand{\method}{GeoSplit}
\newcommand{\E}{\mathbb{E}}
\newcommand{\ind}{\mathbf{1}}
\title{GeoSplit: Selective Edge--Server Map Acquisition for Cross-User Travel-Mode Recognition}
\name{Zepu Wang$^{1,\star}$, Luyao Niu$^{2,\star}$\thanks{$^{\star}$Equal Contribution.}, Ruolin Li$^{2}$, Jeff (Xuegang) Ban$^{1,\dagger}$%
\thanks{$^{\dagger}$Corresponding author.}}
\address{\upshape
$^{1}$University of Washington \quad
         $^{2}$University of Southern California\\
         \texttt{zepu@uw.edu, luyaoniu@usc.edu, ruolinl@usc.edu,
         banx@uw.edu}}

\begin{document}
\maketitle

\begin{abstract}
Networked mobility sensing couples local trajectory inference with remote geospatial context, making communication decisions part of the prediction system. We present \method, an edge--server framework for travel-mode recognition that separates the utility of map evidence, its training-time transfer, and its selective acquisition at runtime. On GeoLife with 10\% labels, we evaluate these components under user-day and user-disjoint grouping. For unseen users, aligned full-map context raises macro-F1 from 0.7154 to 0.7864, while map-informed distillation yields a motion-only student at 0.7225. Through a compact query interface, entropy gating achieves 0.7729 macro-F1 versus 0.7738 for all queries, querying 39.5\% of trajectories and reducing application payload by 60.5\%. Under user-day grouping, selective and all-query scores are similarly close at a 21.6\% query rate. Learned value estimation and probability fusion offer no consistent improvement. These results illustrate communication-efficient edge--server inference for an intelligent transportation sensing task.
\end{abstract}


\section{Introduction}
\label{sec:intro}
Mobile sensing systems infer travel modes from GPS trajectories for passive surveys and mobility analytics~\cite{gpssurvey,travelreview}. Motion features may confuse modes with similar movement patterns~\cite{zheng,dabiri}; geographic context from transit stops and railway lines can help resolve them~\cite{gismode}. When this context resides at a remote map service, recognition quality and communication cost become coupled, especially for participants absent from training. The deployment question is whether shared spatial information is worth its application payload for a particular trajectory.

An edge device can query the server for every trajectory or acquire context selectively. The latter requires estimating query value: local predictive uncertainty need not match the expected reduction in loss from server assistance. Remote context can correct one prediction and harm another. Thus, gains from always-available maps do not establish an efficient selective policy.

Related work studies task-oriented communication~\cite{shao}, cost-aware feature acquisition~\cite{shim}, selective prediction and deferral~\cite{selectivenet,defer}, and distillation from privileged information~\cite{hinton,privileged}. These mechanisms suggest three questions for networked mobility sensing: how useful is aligned remote context, how much can be transferred offline, and when should the device communicate? The last question matters for unseen users because uncertainty estimates can degrade under distribution shift~\cite{uncertainty}.

We study these questions with \method, an edge--server framework that predicts locally from quantized motion features and requests compact map context for selected trajectories. We compare aligned and mismatched maps and motion-only students trained with different teachers, then evaluate query gates against application payload. Pipelines are fitted independently under user-day and user-disjoint grouping. At 10\% labels, aligned maps improve unseen-user macro-F1 from 0.7154 to 0.7864, while a map-informed motion-only student reaches 0.7225. Entropy gating reaches 0.7729 versus 0.7738 for all compact queries while querying 39.5\% of trajectories. This separates the utility of remote evidence from its selective acquisition.

\section{Communication-Aware Map Acquisition}
\label{sec:method}

\begin{figure*}[t]
\centering
\includegraphics[width=\textwidth]{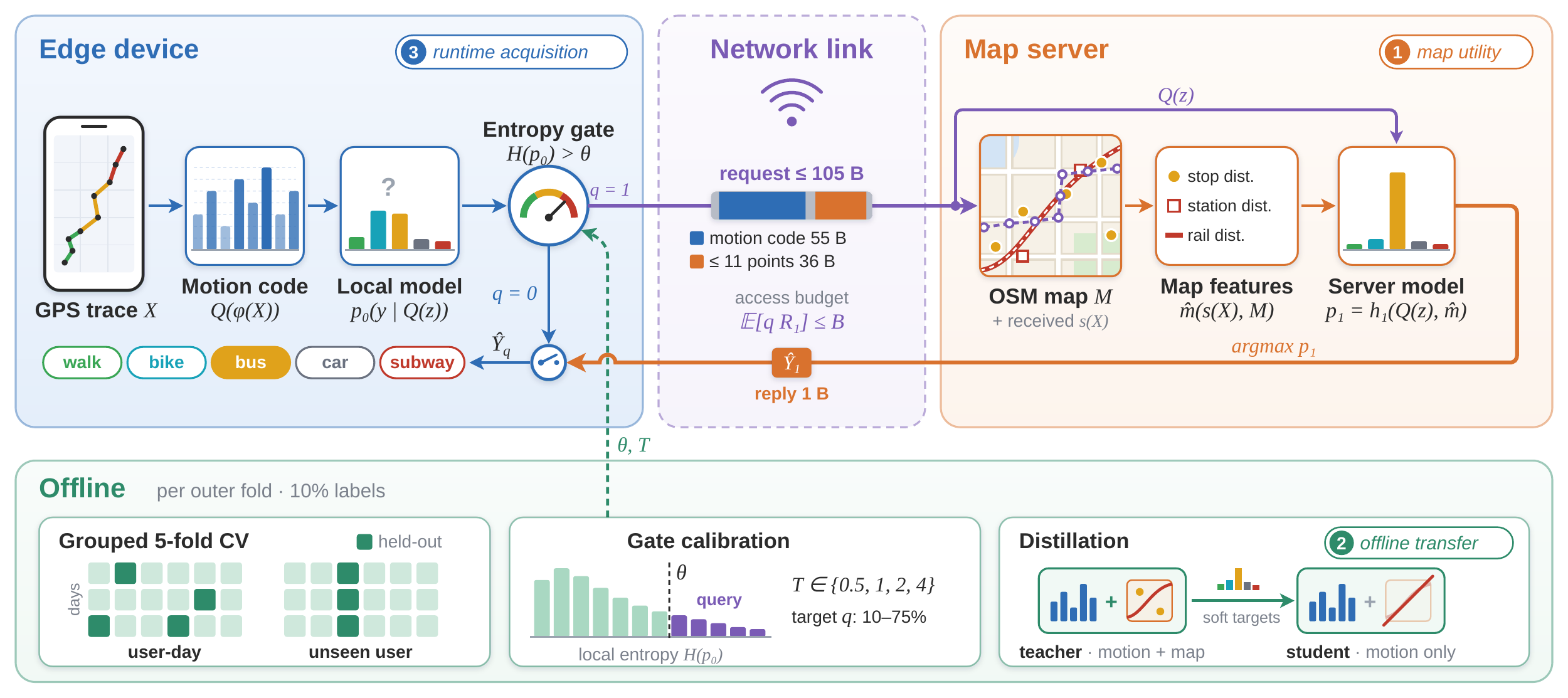}
\caption{\method: local motion prediction with gated map-server assistance. Selected trajectories transmit compact motion and geometry; the server returns a class label. Offline calibration and distillation support evaluation of map utility, training-time transfer, and runtime acquisition.}
\label{fig:framework}
\end{figure*}
\subsection{Local evidence and the query interface}
Figure~\ref{fig:framework} summarizes \method. For a segmented GPS trajectory $X=\{(\ell_t,\tau_t)\}_{t=1}^{T}$ with positions $\ell_t$, timestamps $\tau_t$, and mode $Y\in\{1,\ldots,5\}$, the device extracts motion statistics $z=\phi(X)$ and computes $p_0(y\mid Q(z))$ using a training-fitted quantizer $Q$. A query sends quantized motion and a position summary $s(X)$ to a server holding map $M$:
\begin{equation}
\begin{aligned}
u&=\operatorname{encode}\bigl(Q(z),s(X)\bigr),\\
p_1(y)&=h_1\bigl(Q(z),\widehat m(s(X),M)\bigr).
\end{aligned}
\label{eq:message}
\end{equation}
Here $\widehat m$ extracts map features from transmitted geometry. Both predictors use the same decoded motion; the server adds spatial context and returns a one-byte class, $\widehat Y_1=\arg\max_y p_1(y)$. A gate $q(I)$ using only local information selects $\widehat Y_1$ when $q=1$ and $\widehat Y_0=\arg\max_y p_0(y)$ otherwise, before observing remote features.

\subsection{Query value and selection}
Let $R_1(X)$ count request-plus-reply application bytes, with zero query cost for local inference. We consider
\begin{equation}
\min_q\ \mathcal R(q)=\E[\ell(Y,\widehat Y_q)]
\quad\mathrm{s.t.}\quad
\E[q(I)R_1(X)]\le B.
\label{eq:objective}
\end{equation}
For fixed predictors, conditional query value gives
\begin{equation}
\begin{aligned}
v(I)&=\E[\ell(Y,\widehat Y_0)-\ell(Y,\widehat Y_1)\mid I],\\
\mathcal R(q)&=\mathcal R(0)-\E[q(I)v(I)].
\end{aligned}
\label{eq:value}
\end{equation}
Under zero--one loss, $v(I)$ is the probability of correcting a local error minus that of introducing one. The pointwise Lagrangian rule queries when $v(I)>\lambda\E[R_1(X)\mid I]$~\cite{shim,defer}. We evaluate macro-F1 alongside cost; it is not the additive loss in (\ref{eq:objective}).

The reference gate thresholds local predictive entropy:
\begin{equation}
q(X)=\ind\!\left[-\sum_{c=1}^{5}p_0(c)\log p_0(c)>\theta\right].
\label{eq:gate}
\end{equation}
Alternatives use confidence, margin, random selection, or a learned value score. Thresholds target 10, 25, 50, and 75\% queries within the training partition and remain fixed on outer folds, where realized rates and costs may differ.

The value regressor uses local motion, class probabilities, entropy, margin, and confidence. Its target is local minus blended-query log-loss on labeled out-of-fold predictions, clipped to $[-5,5]$. The blend is $\widetilde p_1=(1-\alpha)p_0+\alpha p_1$, with $\alpha\in\{0,0.25,0.5,0.75,1\}$ selected by calibration log-loss. Gates are compared with both direct and blended predictions. Primary cost comparisons use class-only replies; exploratory fusion requires probabilities and is not assigned this reply cost.

\subsection{Motion and position encoding}
Each of 110 motion statistics uses a four-bit empirical-quantile code: 55 bytes plus a four-byte header. Fine queries encode at most 11 points, selected by maximum-geometric-deviation refinement on a 50\,m grid, in at most 36 bytes plus a seven-byte header. A two-byte frame and one-byte reply give a 105-byte total maximum. Coarse queries use at most four uniformly selected points on a 100\,m grid and 81 bytes overall.

Map features comprise bus-stop and station distances, proximity ratios at 50 and 100\,m, and railway-segment distances. The full-map reference uses original trajectories and adds median stopping-center distances, with a sentinel and missing-context indicator. Classifiers are trained on their respective feature definitions, so full-map and compact-query comparisons reflect both geometry and feature differences. Geometry was selected in an earlier held-out comparison and is fixed here.

\section{Evaluation Design}
\label{sec:experiments}
\subsection{Data and participant protocols}
We classify five GeoLife~\cite{zheng} modes: walk, bike, bus, car (including taxi), and subway. Annotations define segment boundaries. Restricting complete trajectories to Beijing yields 8,731 segments from 64 users. Grouped cross-validation uses only the 6,277-segment, 62-user training pool of an earlier 6,277/695/1,759 training/development/test user-day partition.

Two five-fold outer protocols control participant overlap. \emph{User-day} grouping keeps each user's same-day observations together and merges days linked by crossing trajectories; users can appear in multiple folds. \emph{User-disjoint} grouping holds out entire users~\cite{subjectcv}. Each protocol independently fits models and acquisition rules, so comparisons also reflect differences in routes and sample composition.

A fixed stratified draw labels 10\% of each outer-training pool. Three inner group folds generate out-of-fold predictions, refitting quantizers within each inner-training fold. Labeled groups are split 60/40 for acquisition training and calibration. Calibration selects temperature $T\in\{0.5,1,2,4\}$~\cite{calibration}, fusion coefficients, and query thresholds; outer labels are used only for evaluation. A 100\%-label control repeats the information-source comparisons.

All map comparisons use a Geofabrik OpenStreetMap~\cite{osm} snapshot dated September 19, 2026, buffered around Beijing: 23,930 bus-stop elements, 712 rail/subway stations, and 16,186 railway ways. Thus contemporary maps classify historical trajectories.

\subsection{Spatial and transfer controls}
Spatial controls include map-only prediction, motion plus a centroid on a 1\,km grid, map lookup after a 1\,km eastward trajectory shift, and map features shuffled separately within training and evaluation partitions. Aligned, shifted, and shuffled contexts use matched classifiers.

For offline transfer~\cite{hinton,privileged}, identical motion-only students regress local-teacher or full-map-teacher probabilities, then clip and normalize their outputs. Both use all outer-training trajectories, the same label budget, and cross-fitted teacher targets; labeled samples average these targets with one-hot labels. The teacher input is the only difference. The supervised motion-only predictor provides a further reference.

\subsection{Models, metrics, and communication cost}
Histogram gradient-boosting classifiers use 150 iterations, at most 15 leaves, minimum leaf size 15, learning rate 0.05, and $L_2=1$. The value regressor uses 100 iterations, seven leaves, minimum leaf size 20, and $L_2=10$; students use five regressors with classifier settings. We report pooled outer-fold macro-F1 and paired 95\% intervals from 2,000 user-bootstrap resamples, conditional on fitted models.

Payload includes requests and class replies, assuming shared models and codebooks. Central reporting adds one application byte per trajectory; query totals exclude survey metadata and reporting transport. We also consider fixed per-query overheads of 64 and 256 bytes. Local preloading requires a 2.01\,MB compressed task map and a 16.8\,kB motion codebook. Costs measure application bytes and storage, not wireless latency or energy.

\begin{figure*}[t]
\centering
\includegraphics[width=0.8\textwidth]{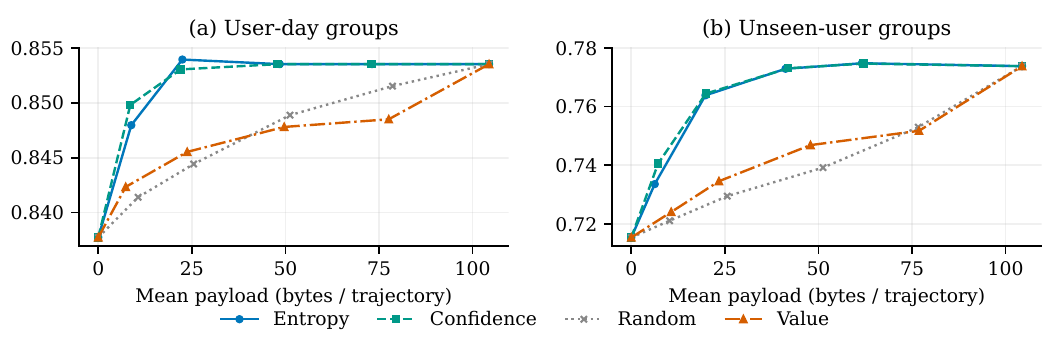}
\caption{Recognition--payload tradeoffs with 10\% labels. Policies share predictors within each independently fitted protocol. Thresholds target 10/25/50/75\% queries on inner calibration data; costs are realized outer-fold payloads. All-query scores use the fine interface.}
\label{fig:frontier}
\end{figure*}

\section{Results}
\label{sec:results}
\subsection{Aligned remote context adds evidence}
Aligned full-map context improves macro-F1 from 0.8377 to 0.8592 under user-day grouping and from 0.7154 to 0.7864 for unseen users (Table~\ref{tab:context}). Paired gains are 0.0215 [0.0119, 0.0438] and 0.0711 [0.0317, 0.1023], respectively; the larger unseen-user gain accompanies a lower motion-only score.

\begin{table}[t]
\centering
\caption{Spatial evidence and training-time transfer with 10\% labels. Entries are pooled outer-fold macro-F1. Both students use only motion at inference.}
\label{tab:context}
\begin{tabular}{lrr}
\toprule
Input or model & User-day & Unseen user\\
\midrule
Motion only & 0.8377 & 0.7154\\
Map only & 0.6726 & 0.5602\\
Motion + coarse location & 0.8404 & 0.7059\\
Motion + full map & \textbf{0.8592} & \textbf{0.7864}\\
Motion + shifted map & 0.8388 & 0.7122\\
Motion + shuffled map & 0.8336 & 0.7142\\
\midrule
Local-teacher student & 0.8264 & 0.6922\\
Map-teacher student & 0.8458 & 0.7225\\
\bottomrule
\end{tabular}
\end{table}

Coarse location does not reproduce this improvement, and shifted or shuffled maps largely remove it, supporting the value of spatial correspondence beyond the tested regional cue. Full-map gains persist at 100\% labels: 0.8843 to 0.9061 under user-day grouping and 0.8051 to 0.8364 for unseen users.

\subsection{Offline transfer leaves a gap to runtime context}
Map-teacher students outperform local-teacher students by 0.0195 and 0.0303 under user-day and user-disjoint grouping. For unseen users, their score of 0.7225 exceeds supervised motion-only prediction by 0.0071 but remains below the full-map score of 0.7864. Thus the evaluated distillation procedure transfers only part of the map benefit, motivating runtime context access.

\subsection{Selective communication approaches all-query performance}
Uncertainty gates improve the observed recognition--payload tradeoff over random queries (Fig.~\ref{fig:frontier}). All fine queries reach 0.8535 under user-day grouping and 0.7738 for unseen users. Selection is compared against these same compact-query predictors.

\begin{table}[t]
\centering
\caption{Observed operating points with 10\% labels. Entropy percentages are calibration targets; $q$ and mean bytes per trajectory are measured on outer folds. Query policies use the fine-query interface.}
\label{tab:query}
\begin{tabular}{llrrr}
\toprule
Grouping & Policy & $q$ (\%) & Bytes & F1\\
\midrule
User-day & Local only & 0.0 & 0.00 & 0.8377\\
 & All fine & 100.0 & 104.39 & 0.8535\\
 & Entropy 25\% & 21.6 & 22.46 & 0.8539\\
 & Entropy 50\% & 46.4 & 48.46 & 0.8535\\
\midrule
Unseen user & Local only & 0.0 & 0.00 & 0.7154\\
 & All fine & 100.0 & 104.39 & 0.7738\\
 & Entropy 25\% & 19.0 & 19.80 & 0.7638\\
 & Entropy 50\% & 39.5 & 41.21 & 0.7729\\
\bottomrule
\end{tabular}
\end{table}

At the 25\% calibration target, user-day entropy gating queries 21.6\% of trajectories and scores 0.8539 versus 0.8535 for all fine queries (Table~\ref{tab:query}). Payload falls from 104.39 to 22.46 bytes per trajectory, a 78.5\% reduction; the paired F1 difference is 0.0004 [$-0.0030$, 0.0024].

For unseen users, the same target yields 19.0\% queries and 0.7638 macro-F1. The 50\% target yields 39.5\% queries and 0.7729 versus 0.7738 for all fine queries, with a paired difference of $-0.0009$ [$-0.0048$, 0.0050]. Payload falls to 41.21 bytes, a 60.5\% reduction. These are observed operating points, not minimum-budget estimates or equivalence guarantees.

All fine queries correct 279 local errors but introduce 135 for unseen users. This supports assessing the value of replacing a prediction, as in (\ref{eq:value}), rather than uncertainty alone.

\subsection{Learned value and fusion offer no consistent advantage}
At the 25\% target, value gating scores 0.8455 and 0.7345 under user-day and user-disjoint grouping, below entropy's 0.8539 and 0.7638, at higher payloads of 23.74 and 23.41 bytes. The ordering persists with matched blended predictions, including the value-target predictor. The evaluated regressor therefore offers no advantage over entropy under this training budget.

Fusion raises user-day entropy macro-F1 from 0.8539 to 0.8563 but lowers unseen-user F1 from 0.7638 to 0.7423. Calibration-log-loss selection thus does not consistently improve outer-fold macro-F1; acquisition and fusion require separate evaluation.

\section{Discussion and Conclusion}
\label{sec:conclusion}
\method\ shows that aligned map evidence improves recognition for unseen users, while offline distillation captures only part of its benefit. Selective compact queries nearly match all-query performance (0.7729 versus 0.7738 macro-F1) at a 39.5\% query rate and 60.5\% lower measured application payload. The user-day setting offers a lower-rate operating point, but participant overlap makes it a different deployment question from the unseen-user evaluation. The observed closeness does not establish equivalence for other participants or maps. These savings assume on-demand map access; a locally preloaded 2.01\,MB map remains an alternative. The evaluation is limited to one dataset, annotation-based segmentation, contemporary maps, and one label draw per fold; bootstrap intervals are conditional on fitted models. Networked mobility sensing should therefore assess remote evidence, offline transfer, and runtime communication together under participant-disjoint evaluation.

\bibliographystyle{IEEEbib}
\bibliography{references}
\end{document}